\documentclass[conference]{IEEEtran}
\IEEEoverridecommandlockouts
\usepackage{cite}
\usepackage{amsmath,amssymb,amsfonts}
\usepackage{algorithmic}
\usepackage{graphicx}
\usepackage{caption}
\usepackage{subcaption}
\usepackage{textcomp}
\usepackage{xcolor}
\def\BibTeX{{\rm B\kern-.05em{\sc i\kern-.025em b}\kern-.08em
    T\kern-.1667em\lower.7ex\hbox{E}\kern-.125emX}}

\usepackage{times} %
\usepackage{soul}
\usepackage{url}
\usepackage{subfig}
\usepackage{float}
\usepackage{amsmath}
\usepackage{amsthm}
\usepackage{booktabs}  %
\usepackage{algorithm}
\usepackage{algorithmic}
\usepackage[switch]{lineno}  %
\usepackage{multirow}
\usepackage[normalem]{ulem}  %
\useunder{\uline}{\ul}{}

\usepackage[american]{babel}  %
\addto\extrasamerican{
}
\newlength\myindent
\begin{document}



\title{Evaluating Generative Models for Tabular Data: Novel Metrics and Benchmarking}


\author{Dayananda Herurkar$^{1,2}$, Ahmad Ali$^{2}$, and Andreas Dengel$^{1,2}$ \\
$^{1}$German Research Center for Artificial Intelligence (DFKI), Kaiserslautern, Germany\\%
$^{2}$RPTU Kaiserslautern-Landau, Germany\\%
email: \texttt{firstname.lastname@dfki/rptu.de}
}

\maketitle

\begin{abstract}

Generative models have revolutionized multiple domains, yet their application to tabular data remains underexplored. 
Evaluating generative models for tabular data presents unique challenges due to structural complexity, large-scale variability, and mixed data types, making it difficult to intuitively capture intricate patterns. 
Existing evaluation metrics offer only partial insights, lacking a comprehensive measure of generative performance. 
To address this limitation, we propose three novel evaluation metrics: FAED, FPCAD, and RFIS. 
Our extensive experimental analysis, conducted on three standard network intrusion detection datasets, compares these metrics with established evaluation methods such as Fidelity, Utility, TSTR, and TRTS. 
Our results demonstrate that FAED effectively captures generative modeling issues overlooked by existing metrics. 
While FPCAD exhibits promising performance, further refinements are necessary to enhance its reliability. 
Our proposed framework provides a robust and practical approach for assessing generative models in tabular data applications.
\end{abstract}


\section{Introduction}

Recent advancements in generative models, particularly diffusion models and transformers, have led to remarkable success across domains such as images, videos, text, and speech. Effective evaluation metrics are critical for assessing model quality and facilitating meaningful comparisons. While qualitative assessments in perceptual domains (e.g., images) rely on human judgment, quantitative metrics like Inception Score (IS) \cite{incep-score} and Fréchet Inception Distance (FID) \cite{fid} have become standard for evaluating generative models in the image domain.

Despite these advancements, there is no widely accepted evaluation standard for generative models applied to tabular data. Existing metrics such as SDV Fidelity, Utility, Privacy through DCR, TSTR \cite{tstr-trts}, and TRTS \cite{tstr-trts} have been employed in specific scenarios; however, their robustness in handling various generative modeling challenges remains insufficiently explored. This inconsistency hinders progress in deep generative models for tabular data and complicates model benchmarking.

Inspired by IS and FID, we introduce three novel evaluation metrics, FAED, FPCAD, and RFIS, tailored for assessing generative models in tabular data. We investigate whether existing evaluation principles from the image domain can be effectively adapted for tabular data. Prior studies have suggested assessment metrics such as SDV Fidelity, Utility, TSTR, and TRTS, yet these methods exhibit limitations in detecting key generative modeling challenges. Our study rigorously evaluates these metrics alongside our newly introduced approaches.

To systematically assess the reliability of evaluation metrics, we introduce three significant challenges into real datasets: Quality Decrease, Mode Drop, and Mode Collapse. We examine whether FAED, FPCAD, and RFIS effectively capture these issues compared to existing metrics. Experimental results on three standard network intrusion detection datasets demonstrate that FAED successfully detects all synthesized problems, whereas FPCAD shows promising but improvable performance. In contrast, existing metrics fail to identify key issues, underscoring the need for more robust evaluation measures for tabular data generative models.

Our study makes the following key contributions:
\begin{itemize}
    \item We introduce FAED and FPCAD, novel and robust metrics for evaluating generative models in the tabular data domain.
    \item We simulate real-world generative modeling challenges by embedding Quality Decrease, Mode Drop, and Mode Collapse issues into the datasets.
    \item We perform an in-depth comparative analysis of existing and proposed metrics, identifying their limitations and advantages in capturing generative modeling issues.
\end{itemize}

By addressing the shortcomings of current evaluation approaches, our work provides a more reliable framework for assessing generative models in tabular data applications.

\section{Related Work}
Quantitatively evaluating the effectiveness of generative models remains a significant challenge \cite{borji2019pros}. 
Since the model's learned distribution is not explicitly accessible, direct validation becomes impractical. 
Additionally, manually verifying the authenticity of synthetic data \cite{herurkar2024tab} is infeasible, especially for complex and non-intuitive formats such as tabular data \cite{herurkar2024fin, herurkar2023recol}. 
As a result, the use of generative models for producing artificial tabular data remains constrained. 
In this section, we list generative approaches for tabular data and examine papers covering methodologies for assessing generative models in the image and time series domain.

\subsection{Generative Approaches for Tabular Data}
This section reviews key generative models designed for synthesizing tabular data, including advancements in GANs, VAEs, and diffusion models. 
Among GAN-based approaches, TableGAN \cite{park2018} employs convolutional neural networks and a built-in classifier to enhance the semantic integrity of generated records while maintaining data privacy. 
TGAN \cite{xu-veer2018} introduces an LSTM-based generator and Gaussian Mixture Models (GMMs) for handling mixed data types, while CTGAN \cite{xu2019} improves upon TGAN by incorporating a Variational Gaussian Mixture Model (VGM) and Wasserstein loss for better feature representation and conditional data generation. 
CopulaGAN \cite{patki2016} extends CTGAN by leveraging copula functions to capture feature dependencies.
Beyond GANs, MedGAN \cite{choi2018} employs an autoencoder-assisted GAN to generate high-dimensional discrete data, particularly for healthcare applications. 
However, it struggles with mixed data types and lacks correlation between separately generated features. 
Variational Autoencoders (VAEs) offer alternative solutions, with TVAE \cite{xu2019} adapting VAE architectures for tabular data, and VAEM \cite{ma2020vaem} improving feature dependency modeling through a two-stage probabilistic approach, making it effective for handling missing values.
More recently, diffusion models have emerged as state-of-the-art generative techniques for tabular data. 
TabDDPM \cite{tabddpm} applies denoising diffusion probabilistic modeling (DDPM) with Gaussian quantile transformations for numerical features and multinomial diffusion for categorical data. 
FinDiff \cite{findiff} tailors diffusion-based generation for financial data, embedding categorical features into continuous vector spaces to capture semantic relationships more effectively than traditional one-hot encoding.
These generative models collectively represent significant progress in synthetic tabular data generation, addressing challenges such as feature dependencies, mixed data types, and data privacy while advancing state-of-the-art performance in various application domains.

\subsection{Qualitative Assessment of Generative Models (Images and Time Series)}
This section gives an overview of the different models used for generating synthetic data on images and times series. In the image domain, GANs have led to significant breakthroughs in generating high-resolution, photorealistic images \cite{karras2018}. However, more recently, diffusion models have emerged as state-of-the-art generative techniques, outperforming GANs in image synthesis quality and diversity. DDPM \cite{ho2020} demonstrated the power of denoising diffusion probabilistic modeling, achieving near-photorealistic synthesis. Imagen\cite{saharia2022} and DALL·E 2 \cite{ramesh2022} leveraged diffusion-based transformers for text-to-image generation, producing high-resolution, contextually accurate images. Stable Diffusion \cite{rombach2022} further optimized efficiency by introducing latent-space diffusion, enabling real-time image synthesis. Time-series generation poses unique challenges due to temporal dependencies and variable-length sequences. Generative models in this domain leverage architectures such as GANs, VAEs, and transformers to model time-evolving patterns. GAN-based approaches have proven effective for generating synthetic time-series data with realistic temporal dependencies. TimeGAN \cite{yoon2019} integrates an autoencoder with recurrent GANs, combining supervised and unsupervised objectives to generate temporally coherent sequences. TTS-GAN \cite{esteban2017} applies GANs to medical time-series data, ensuring realistic patient records while preserving privacy. More recently, diffusion models have been adapted for time-series synthesis. For example, TSDiff \cite{kollovieh2023} applies denoising diffusion modeling, achieving state-of-the-art performance in generating high-fidelity synthetic time-series data while maintaining accurate long-term dependencies.

\section{Quantitative Evaluation}

This section outlines an evaluation pipeline specifically designed for assessing generative models in the tabular data domain. To establish a standardized and reliable framework, we introduce three novel evaluation metrics: FAED, FPCAD, and RFIS, which are visually represented in Figure \ref{fig:gen_eval_metrics}. These metrics, alongside existing evaluation methods, enable a comprehensive assessment of generated samples based on predictive performance. Inspired by evaluation principles from the image domain, FAED and FPCAD leverage statistical distance measures similar to the Fréchet Inception Distance (FID), while RFIS is influenced by the Inception Score. The following subsections provide a detailed description of each metric.

\subsection{Fréchet AutoEncoder Distance (FAED)}
The Fréchet Inception Distance (FID) \cite{fid} is a widely used metric for assessing the quality of generated images by measuring the similarity between real and generated data distributions. Inspired by this approach, we introduce the Fréchet AutoEncoder Distance (FAED) for evaluating generative models in tabular data. FAED is based on Autoencoder (AE) \cite{Herurkar-ijcnn23, sattarov2022explaining} architectures, which encode input data into a latent space and then reconstruct the output to closely resemble the original input. These latent vectors capture essential structural features in a lower-dimensional representation. To compute FAED, we extract feature vectors from the latent layer of a pre-trained AE for both real and generated samples. Assuming these feature representations follow a continuous multivariate Gaussian distribution, we compute the Fréchet Distance (Wasserstein-2 distance) using their mean and covariance matrices represented by below equation

\begin{equation}
    \|\mu_r - \mu_s\|^2 + \text{Tr} \left( \Sigma_r + \Sigma_s - 2 (\Sigma_r \Sigma_s)^{1/2} \right)   \label{eq: fid}
\end{equation}

The steps are shown in Figure\ref{fig:gen_eval_metrics} (a). A lower FAED value signifies a closer match between real and generated data distributions, with an optimal score of zero indicating perfect alignment. FAED serves as a robust and computationally efficient metric, although its effectiveness depends on the assumption that the feature space follows a multivariate Gaussian distribution, which may not always hold across diverse datasets.

\subsection{Fréchet PCA Distance (FPCAD)}
\begin{figure*}[t!]
\centering
\includegraphics[width=\linewidth]{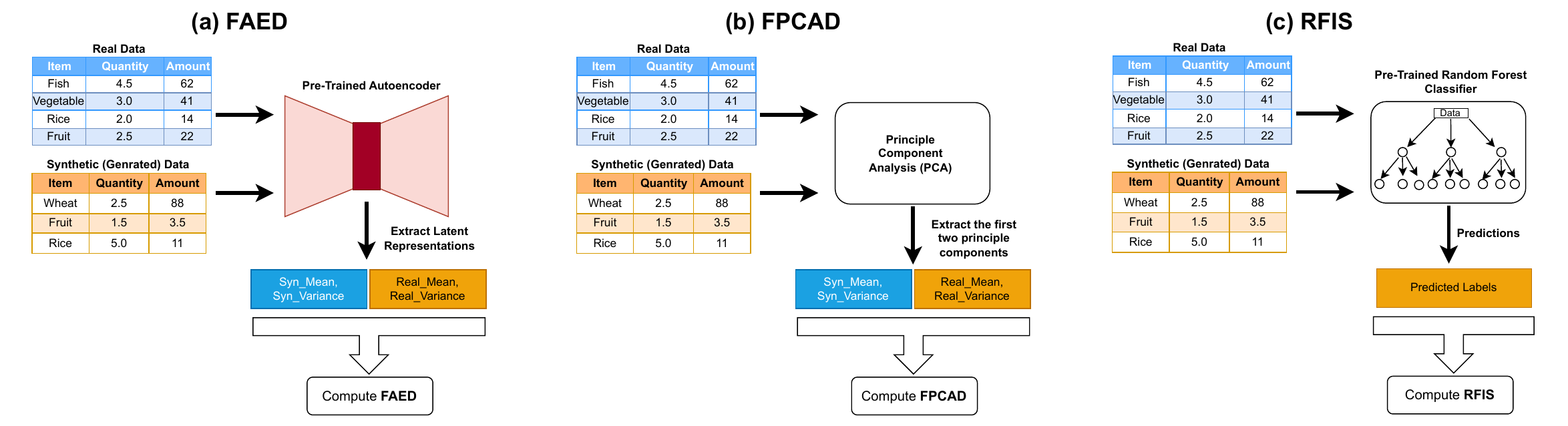}
\caption{Overview of the proposed evaluation metrics for generative models in the tabular data domain. (a) Fréchet AutoEncoder Distance (FAED) computes the similarity between real and generated data distributions using feature representations extracted from a pre-trained autoencoder. (b) Fréchet PCA Distance (FPCAD) follows a similar approach but utilizes PCA-transformed representations instead of autoencoder embeddings, eliminating the need for pre-training. (c) Random Forest Inception Score (RFIS) assesses the quality of generated samples based on entropy differences between conditional and marginal label distributions, drawing inspiration from the Inception Score used in image evaluation.}
\label{fig:gen_eval_metrics}
\end{figure*}

FPCAD follows a methodology similar to FAED but replaces Autoencoders with Principal Component Analysis (PCA). One limitation of FAED is the requirement for a pre-trained AE model, which may introduce dependencies on model-specific variations. To address this, FPCAD provides an alternative that eliminates the need for pre-training. We apply PCA to both real and generated datasets, reducing their dimensionality to principal components that preserve the most significant variance. We then compute the mean and covariance matrices from these transformed data representations and use them to calculate the Fréchet Distance represented by equation \ref{eq: fid} and the complete process is shown in Figure\ref{fig:gen_eval_metrics}(b). Similar to FAED, a lower FPCAD score indicates greater similarity between real and synthetic distributions. By leveraging an online, data-driven transformation, FPCAD presents a lightweight and adaptable evaluation metric that can be applied without additional pre-training overhead.

\subsection{Random Forest Inception Score (RFIS)}
Building on the concept of the Inception Score (IS) \cite{incep-score} used in image domain evaluations, we propose the Random Forest Inception Score (RFIS) for assessing synthetic tabular data. RFIS measures the relationship between generated samples and their predictive consistency, evaluating the entropy difference between conditional and marginal label distributions. Let \textit{x} represent the generated samples and \textit{y} their corresponding labels. High-quality synthetic data should exhibit a conditional label distribution $p(y|x)$ with low entropy, while the marginal distribution \textit{p(y)} should be relatively uniform, indicating diversity. The RFIS formulation calculates the difference between these entropy measures: 
\begin{equation}
    RFIS = E[H(p(y)) - H(p(y|x))]
\end{equation}
where \textit{H(.)} represents entropy. A higher RFIS score indicates better generative performance, as it captures the contrast between structured conditional distributions and diverse marginal distributions. FAED and FPCAD both rely on both generated and real samples whereas RFIS relies solely on the statistics of the generated samples and ignores real samples.

\subsection{Other Evaluation Metrics}
\textbf{SDV Fidelity: } The SDV framework \cite{sdmetrics} provides a systematic approach for assessing synthetic data quality by comparing it to real data across multiple statistical and structural dimensions. Fidelity is a key measure, evaluating the degree of resemblance between synthetic and real datasets at both the column and row levels.
\begin{itemize}
    \item \textbf{Column Fidelity:} For numerical attributes, we use the Kolmogorov-Smirnov (KS) statistic to quantify distributional differences. For categorical attributes, the Total Variation Distance (TVD) is used to compare frequency distributions.
    \begin{equation}
    \omega_{\text{col}}(x^d, s^d) =
    \begin{cases}
        1 - KS(x^d, s^d), & \text{if } d \text{ is numerical,} \\
        1 - \frac{1}{2} TVD(x^d, s^d), & \text{if } d \text{ is categorical.}
    \end{cases}
    \end{equation}
    
    \item \textbf{Row Fidelity:} This metric examines the relationships between attribute pairs. Pearson correlation coefficients are used for numerical attributes, while contingency tables and TVD are applied for categorical attributes.
    \begin{equation}
    \omega_{\text{row}}(x^{a,b}, s^{a,b}) =
    \begin{cases}
        1 - \frac{1}{2} \left| \rho(x^{a,b}) - \rho(s^{a,b}) \right|, & \text{if } a,b \text{ are num,} \\
        1 - \frac{1}{2} TVD(x^{a,b}, s^{a,b}), & \text{if } a,b \text{ are cat.}
    \end{cases}
    \end{equation}
\end{itemize}
An overall fidelity score is computed by averaging column- and row-level evaluations, providing a holistic measure of data realism. \\
\begin{equation}
    SDV Fidelity = 0.5 \, \Omega_{\text{col}}(X, S) + 0.5 \, \Omega_{\text{row}}(X, S)
\end{equation}

\textbf{Utility: } Utility measures the effectiveness of synthetic data in real-world applications, often referred to as Machine Learning efficacy. A fundamental utility test involves training a model on synthetic data and evaluating its performance on real data. If the synthetic dataset captures meaningful structures, models trained on it should generalize well to real data. To quantify utility, we conduct two classifier-based evaluations using a Random Forest model:
\begin{itemize}
    \item \textbf{Train on Synthetic, Test on Real (TSTR):} A classifier is trained on synthetic data and tested on real data. A high TSTR accuracy suggests that synthetic data effectively approximates real-world distributions.
    \item \textbf{Train on Real, Test on Synthetic (TRTS):} A classifier is trained on real data and tested on synthetic data. A high TRTS score indicates that the synthetic data retains key characteristics of the real data.
\end{itemize}
TSTR is particularly useful for detecting cases where synthetic data only partially represents real data, while TRTS assesses whether synthetic samples introduce patterns absent in real data. Together, these measures provide a comprehensive view of synthetic data quality. \\
Through this quantitative evaluation framework, our proposed metrics FAED, FPCAD, and RFIS offer robust tools for assessing generative models in the tabular data domain. By benchmarking against existing evaluation approaches, we establish a systematic methodology for ensuring consistency and reliability in synthetic data assessment.

\section{Experimental Setup}
\subsection{Datasets}
To evaluate the effectiveness of our proposed metrics, we utilize three widely recognized tabular datasets for intrusion detection:
\begin{itemize}
    \item UNSW-NB15: Developed by the Australian Centre for Cybersecurity (ACCS) at UNSW Canberra, this dataset provides a contemporary benchmark for testing intrusion detection systems (IDS).
    \item TON-IoT: The Testbed of Things IoT (TON-IoT) dataset is designed to assess the security of IoT environments, capturing diverse cyberattack scenarios.
    \item CICIDS-2017: Created by the Canadian Institute for Cybersecurity, this dataset simulates real-world benign and malicious network traffic, making it a valuable resource for evaluating IDS.
\end{itemize}
A detailed summary of these datasets, including the number of samples, features, and class distributions, is provided in Table \ref{tab:ds_table}.

\begin{table}[]
\begin{tabular}{@{}lrrrrl@{}}
\toprule
\multicolumn{1}{c}{\multirow{2}{*}{\textbf{Dataset}}} & \multicolumn{1}{c}{\multirow{2}{*}{\textbf{Samples}}} & \multicolumn{3}{c}{\textbf{Columns}} & \multicolumn{1}{c}{\multirow{2}{*}{\textbf{Classes}}} \\ \cmidrule(lr){3-5}
\multicolumn{1}{c}{} & \multicolumn{1}{c}{} & \multicolumn{1}{c}{\textbf{Categ.}} & \multicolumn{1}{c}{\textbf{Numerical}} & \multicolumn{1}{c}{\textbf{Encoded}} & \multicolumn{1}{c}{} \\ \midrule
UNSW-NB15 & 257675 & 5 & 39 & 205 & Binary \\
TON-IOT & 211045 & 30 & 14 & 1632 & Binary \\
CICIDS-2017 & 225714 & 10 & 50 & 4248 & Binary \\ \bottomrule
\end{tabular}
\caption{Description of Datasets}
\label{tab:ds_table}
\end{table}

\subsection{Problem Setup} \label{sec: prob_setup}
Assessing the reliability of an evaluation metric is inherently more complex than evaluating a generative model itself. A dependable metric should not only measure the quality of generated data but also effectively detect fundamental issues in generative modeling. To systematically validate our proposed metrics, we introduce three common generative modeling challenges:
\begin{itemize}
    \item \textbf{Quality Decrease:} Quality decrease refers to the deterioration of the generated data compared to the original data. To test quality decrease we introduced two different noise insertion approaches.
    \begin{itemize}
        \item \textbf{Incremental Noise Level:} This approach involves inserting different levels of noise sampled from a Gaussian distribution into the test data and testing how the different metrics perform in detecting different noise levels. The noise has a mean ($\mu$) of 0, and a standard deviation ($\alpha$), that ranges from 0.1 up to 0.5, with increments of 0.1 every time. This noise is applied to the dataset as a whole, which means that the entire dataset is disrupted once the noise is inserted.
        \item \textbf{Incremental Noisy Rows Percentage:} This approach uses the same noise level ($\alpha$ = 0.5), and inserts noise to the rows in the dataset incrementally, instead of inserting noise to the dataset as a whole. This avoids disrupting the entire dataset at once, and instead, there is a gradual insertion of noise that shows how the metrics perform when the dataset is only partially noisy.
    \end{itemize}
    \item \textbf{Mode Drop:} Mode drop happens when the generative model fails to generate all of the distinct modes present in the real data. A mode in the context of tabular data refers to a region of the feature space that represents a certain subset or grouping of the data with specific characteristics. Mode drop occurs when the model generates only a subset of these modes, essentially missing out on important variations in the data. In order to mimic this problem we tried three different types:
    \begin{itemize}
        \item \textbf{Single Mode Drop:} 0/1 labels were dropped to simulate data without inliers/outliers respectively.
        \item \textbf{Successive Mode Drop:} To simulate a mode drop for our datasets, the most repeated combinations of values in the data were ranked from top to bottom and the top 5 were dropped successively. This mimics a very common condition that occurs when a generative model is unable to capture some modes within the dataset during training.
        \item \textbf{Extreme Mode Drop:} The least used combinations from the same list used in the successive mode drop were kept successively in different amounts (\% of the total number of combinations in the dataset), while the rest was dropped. This is a more extreme condition where only the least common types of internet traffic were captured.
    \end{itemize}
    \item \textbf{Mode Collapse:} Mode collapse occurs when the generative model only learns to generate data from one or a few of the modes of the original distribution, even though the real data has multiple modes. As a result, the model generates data that is not diverse and fails to capture the full variety of real data. In the context of tabular data, the model might consistently generate the same values or combinations of values for certain features, while other realistic possibilities are not represented at all.
    \begin{itemize}
        \item \textbf{Label Split:} The dataset was split based on the 0/1 label, then for each of the 2 groups the rows were replaced by the most repeated categorical values and the mean for the numerical values in each of the 2 groups.
        \item \textbf{No Label Split:} The rows were replaced by the most repeated categorical values and the mean for the numerical values of the entire dataset, then the labels were filled as 0/1 afterward in a random fashion.
    \end{itemize}
\end{itemize}

\subsection{Computing Evaluation Score}
To assess the effectiveness of our proposed metrics (FAED, FPCAD, RFIS) in detecting these generative modeling issues, we employ the following experimental protocol:
\begin{enumerate}
    \item Split each dataset into train-test set.
    \item Treat the train set as it is and take test set as the target set.
    \item By applying all the evaluation metrics compute baseline scores ($score_{base}$) i.e, the score obtained by comparing test data against the train set, serving as the highest achievable reference value.
    \item Introduce controlled distortions in test data as described in \ref{sec: prob_setup} and evaluate the response of each metric by comparing the distorted test data against the train set as $score_{gen}$.
\end{enumerate}
The relative performance of the evaluation metrics is quantified by comparing $score_{gen}$ with $score_{base}$ represented as 
\begin{equation}
    rel(score) = score_{base} - score_{gen}
\end{equation} 
A robust metric should demonstrate sensitivity to all induced issues while maintaining stability under ideal conditions. The results of these experiments provide insight into the effectiveness of FAED, FPCAD, and RFIS in assessing the quality and reliability of generative models for tabular data.

\section{Result}
In this section we discuss the result of three experiments. Code available\footnote{Code available: \url{https://github.com/madhajj/Thesis_Ali}}
\subsection{Experiment1: Quality Decrease}

\begin{figure}[h!] 
    \centering
    \begin{subfigure}{0.48\columnwidth}
        \includegraphics[width=\linewidth]{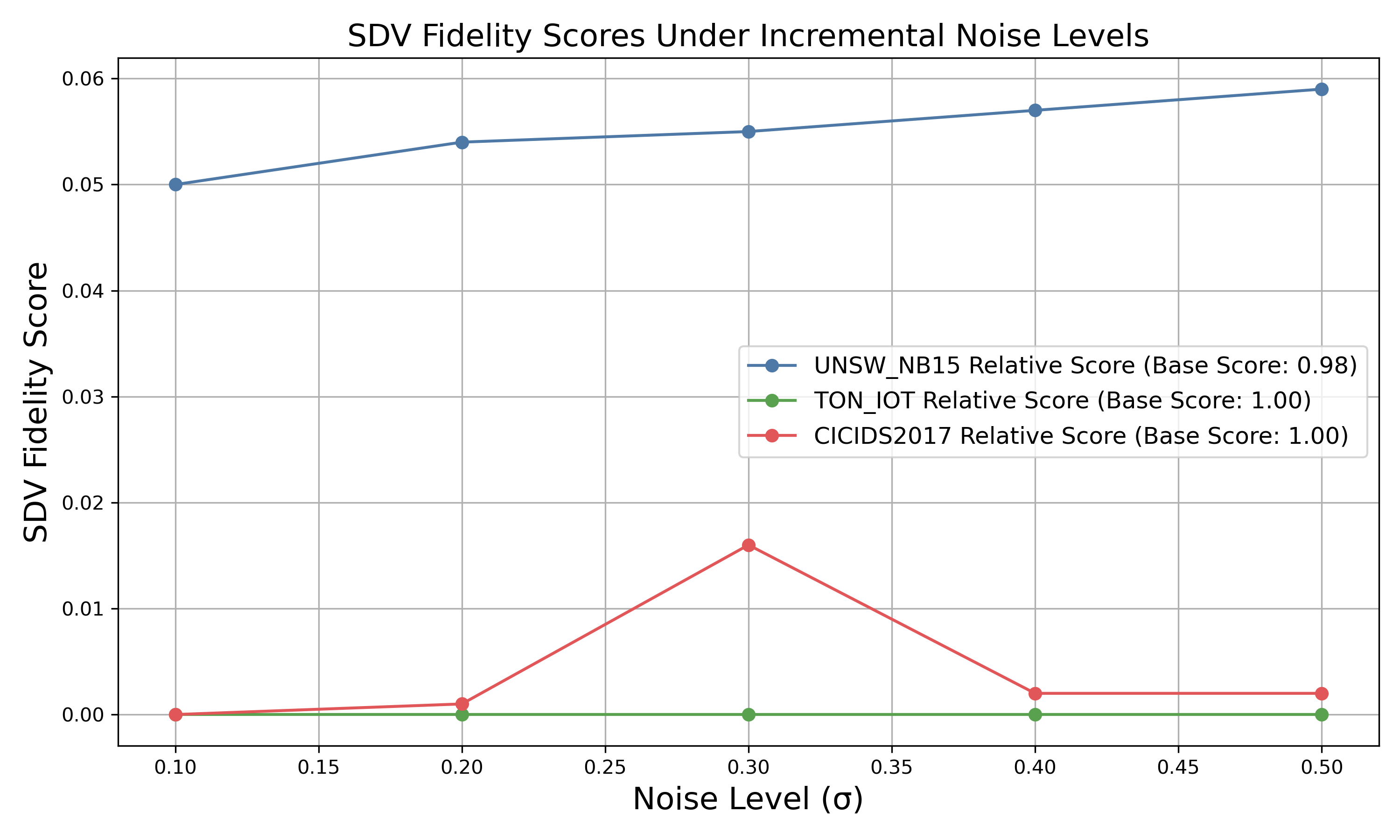}
    \end{subfigure}
    \begin{subfigure}{0.48\columnwidth}
        \includegraphics[width=\linewidth]{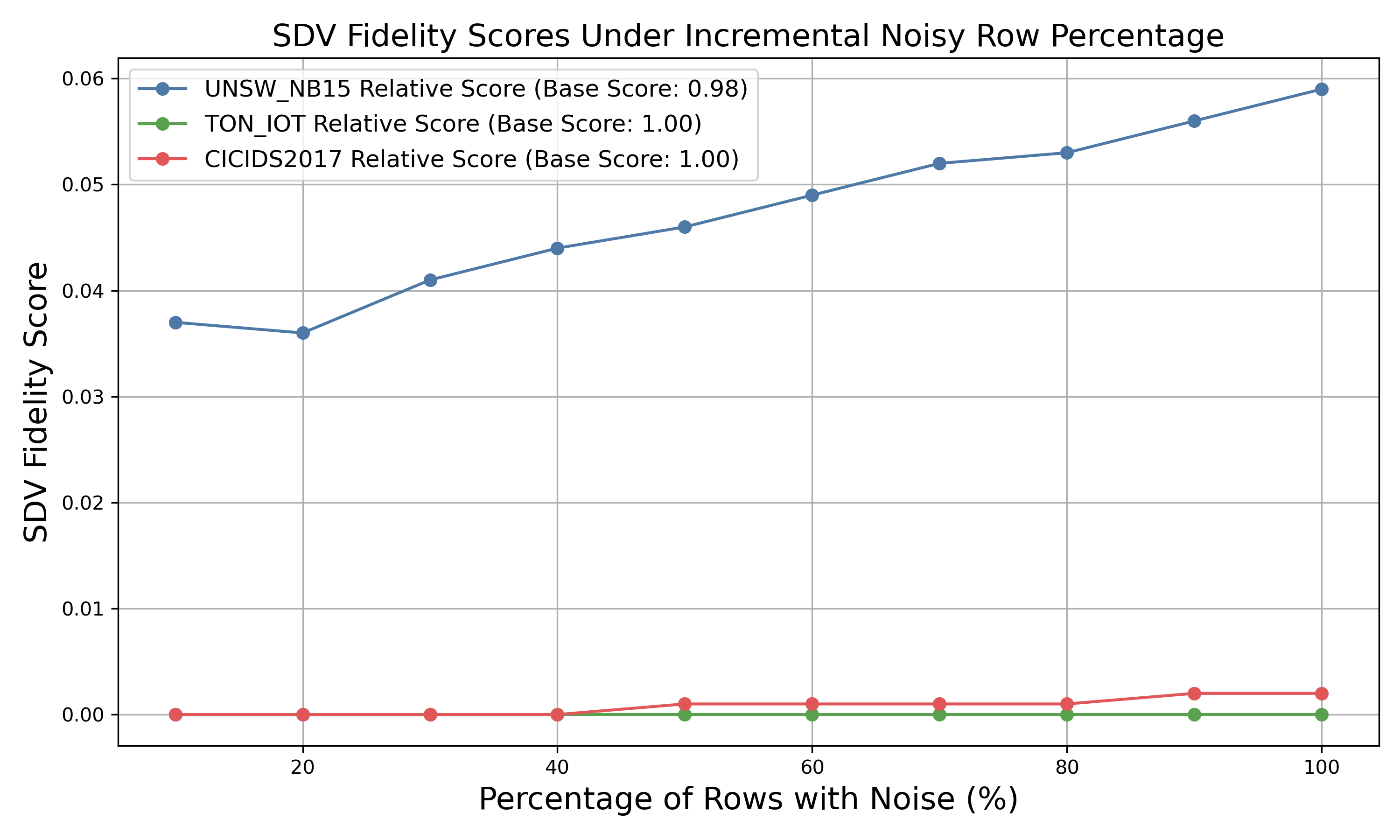}
    \end{subfigure}
    
    \begin{subfigure}{0.48\columnwidth}
        \includegraphics[width=\linewidth]{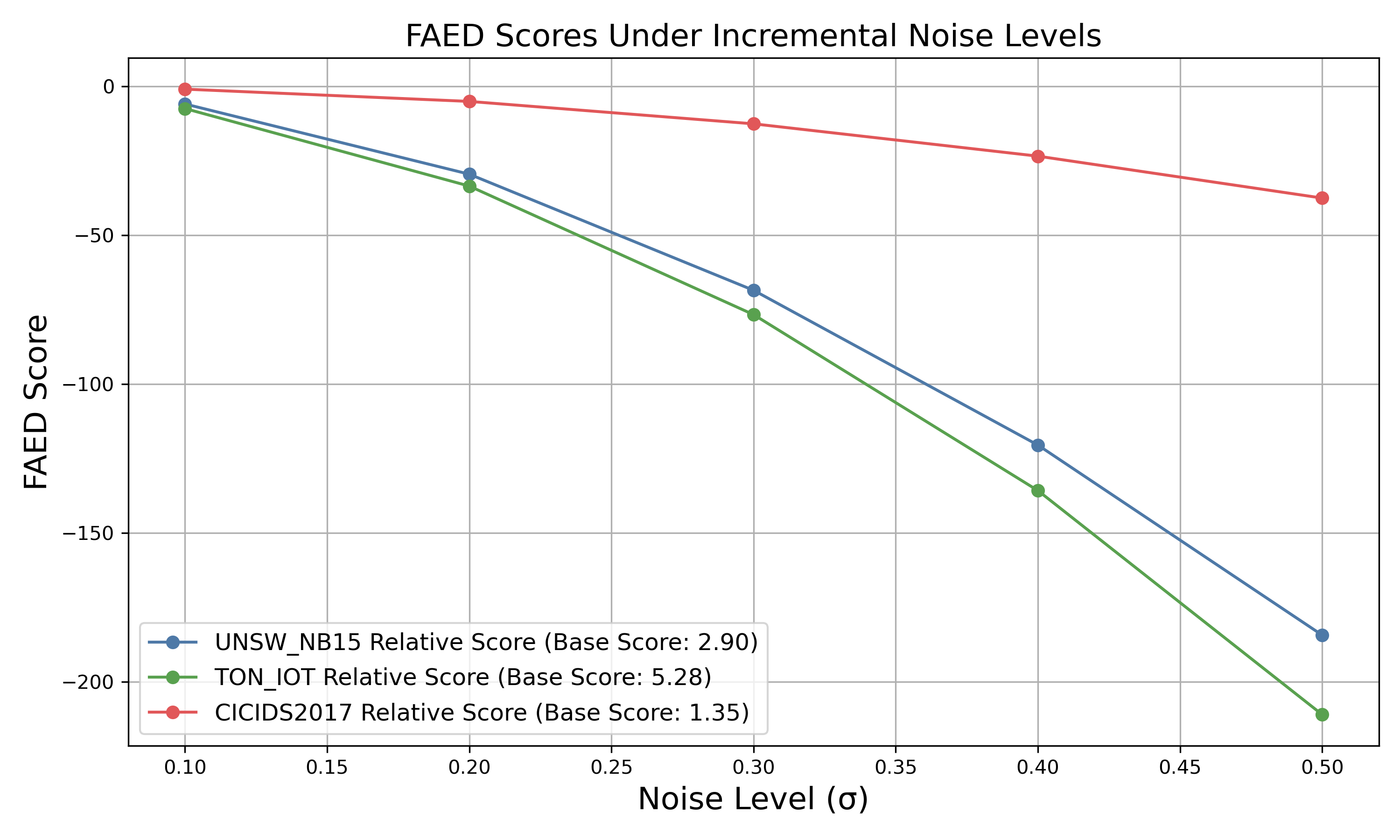}
    \end{subfigure}
    \begin{subfigure}{0.48\columnwidth}
        \includegraphics[width=\linewidth]{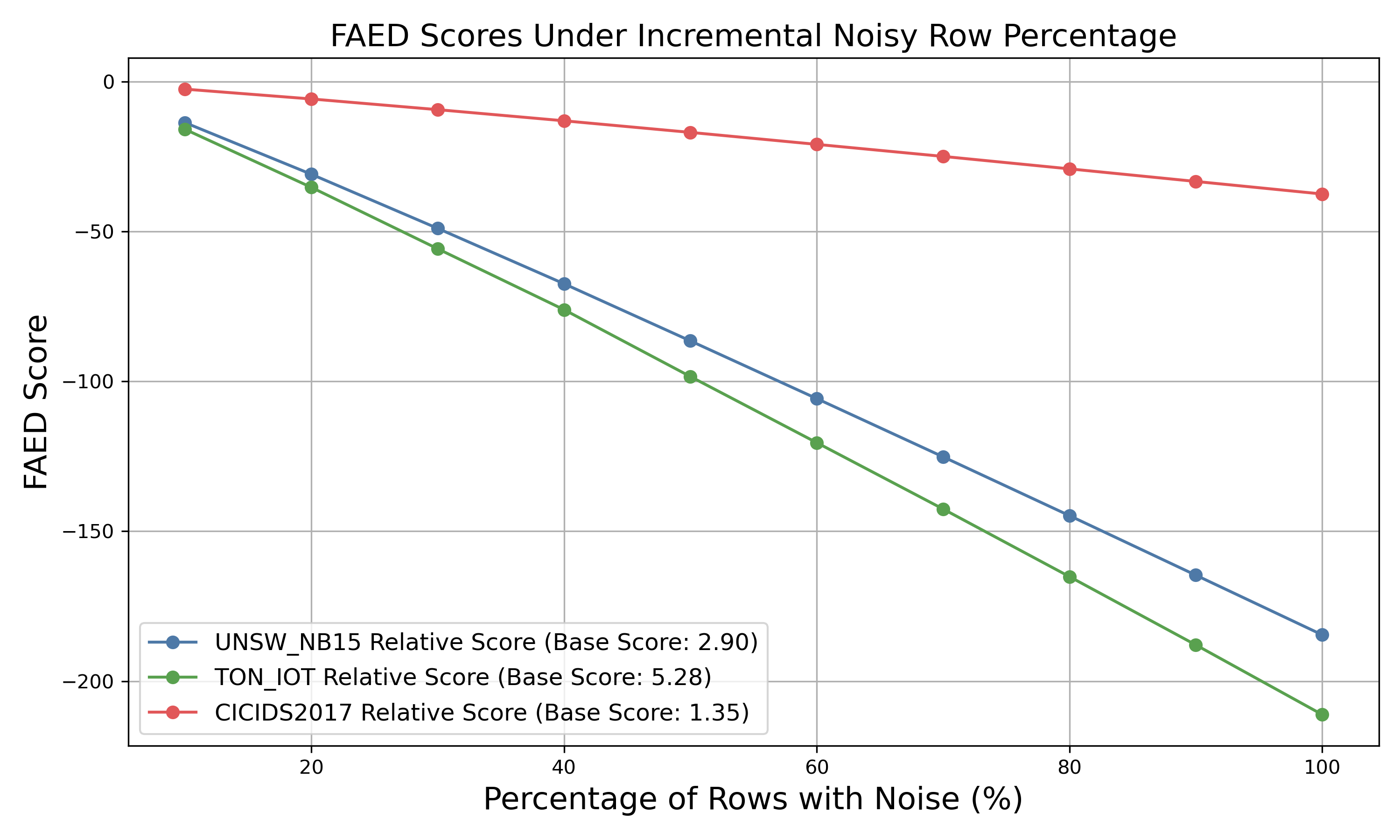}
    \end{subfigure}

    \begin{subfigure}{0.48\columnwidth}
        \includegraphics[width=\linewidth]{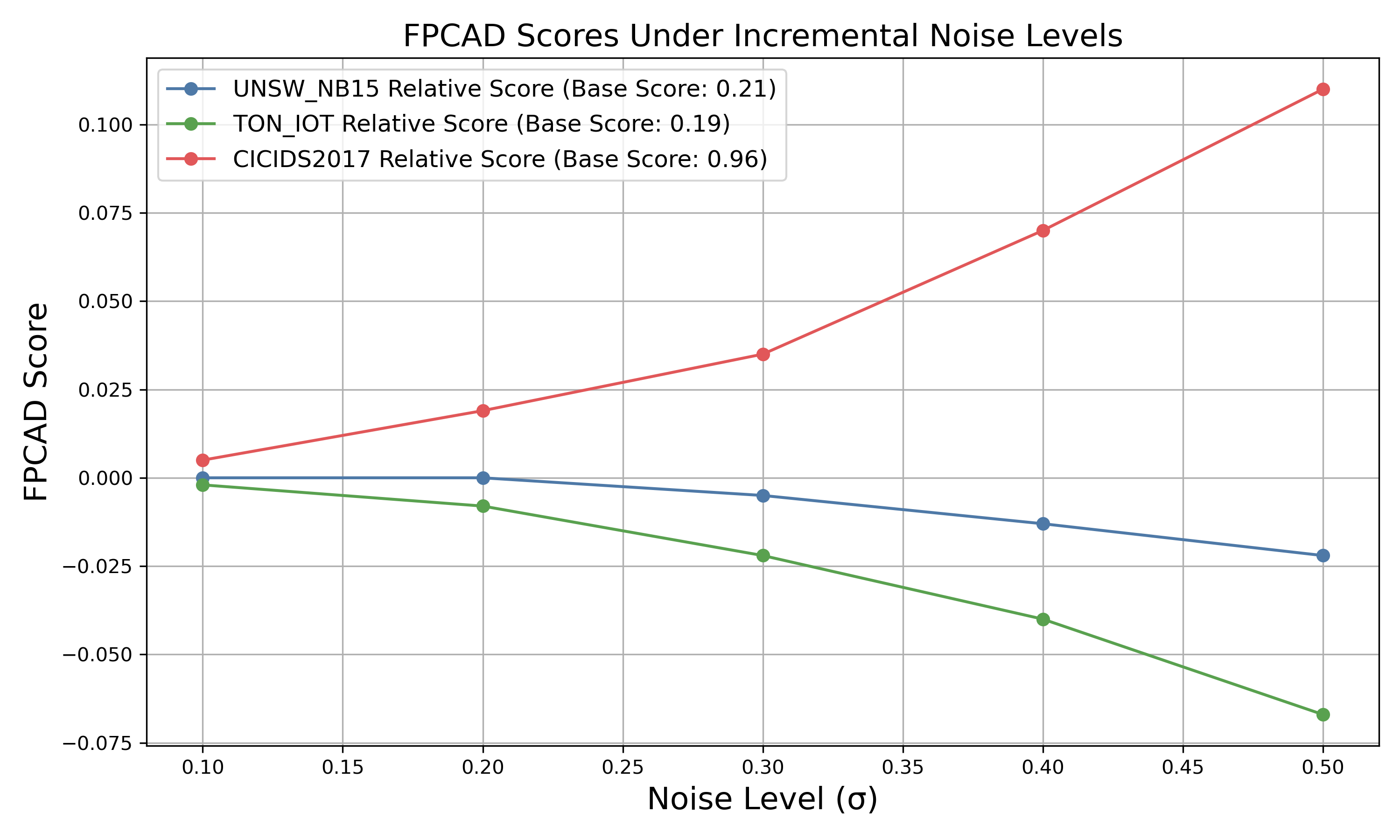}
    \end{subfigure}
    \begin{subfigure}{0.48\columnwidth}
        \includegraphics[width=\linewidth]{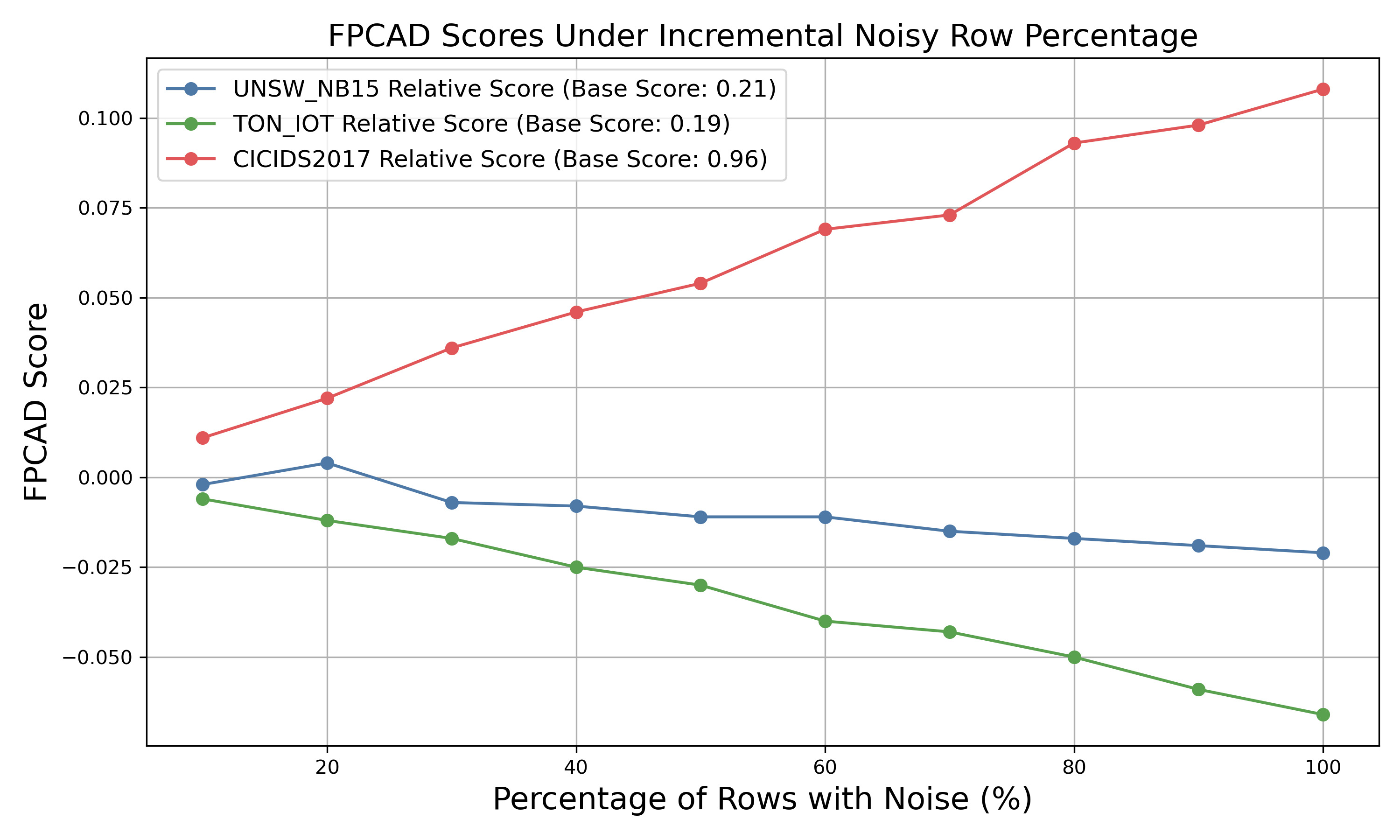}
    \end{subfigure}

    \begin{subfigure}{0.48\columnwidth}
        \includegraphics[width=\linewidth]{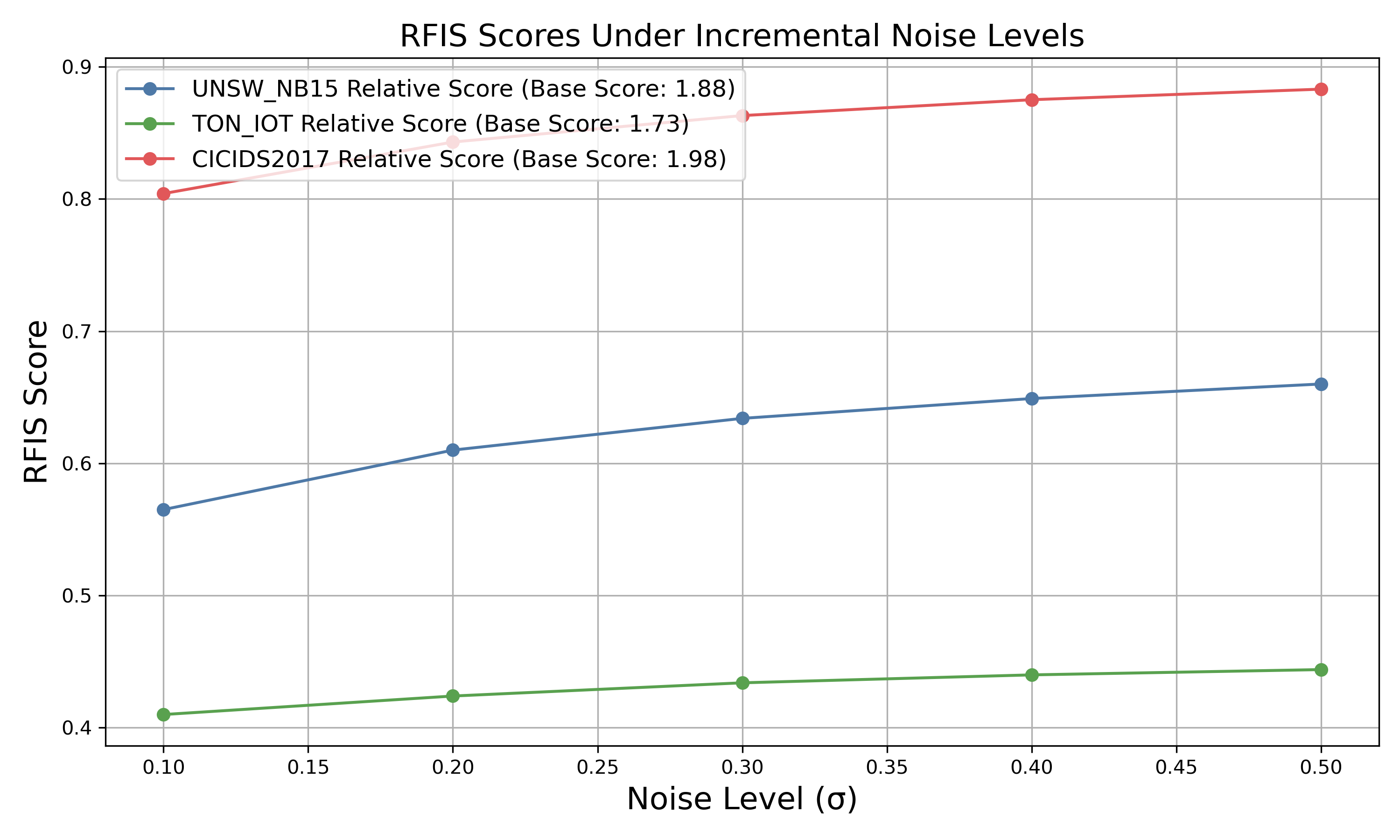}
    \end{subfigure}
    \begin{subfigure}{0.48\columnwidth}
        \includegraphics[width=\linewidth]{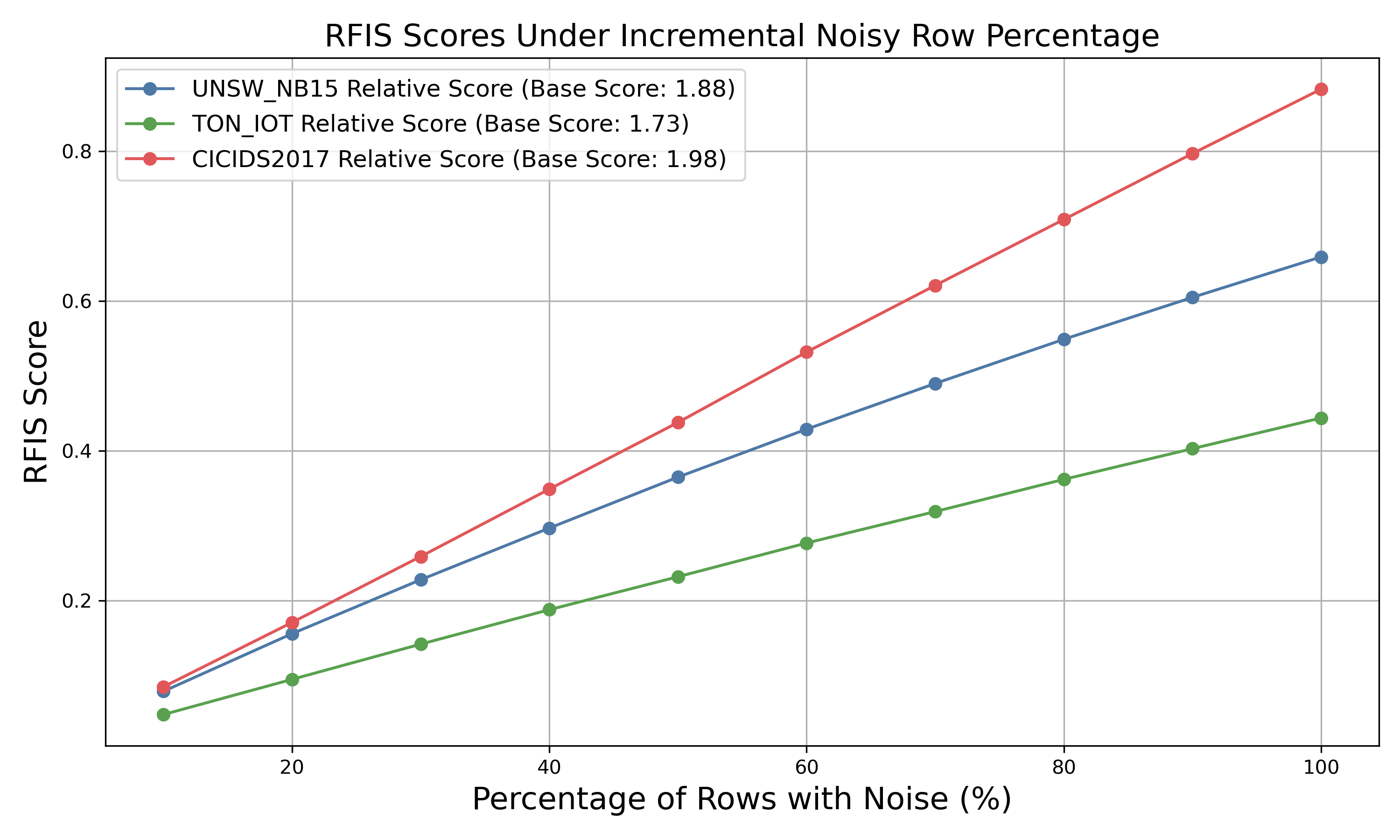}
    \end{subfigure}

    \begin{subfigure}{0.48\columnwidth}
        \includegraphics[width=\linewidth]{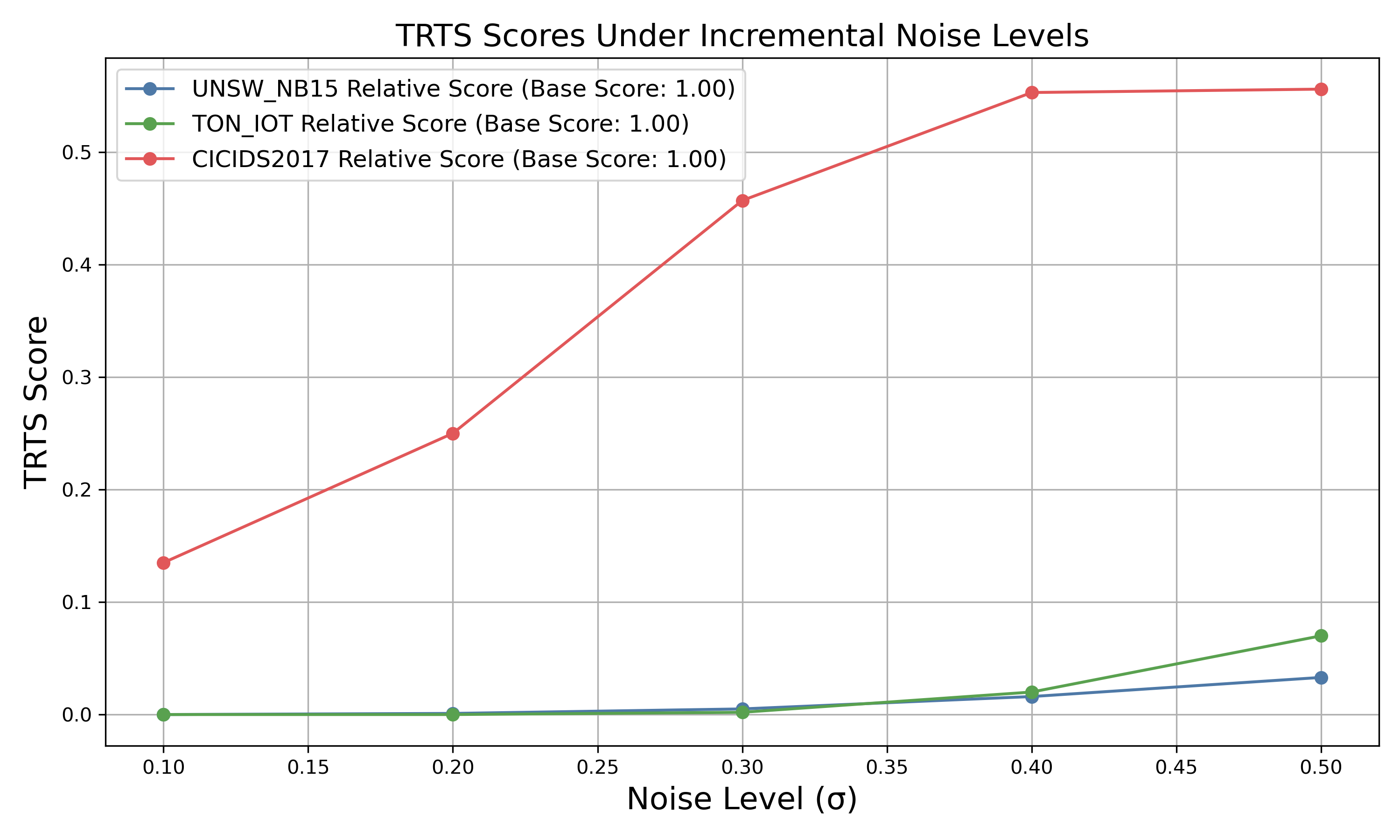}
    \end{subfigure}
    \begin{subfigure}{0.48\columnwidth}
        \includegraphics[width=\linewidth]{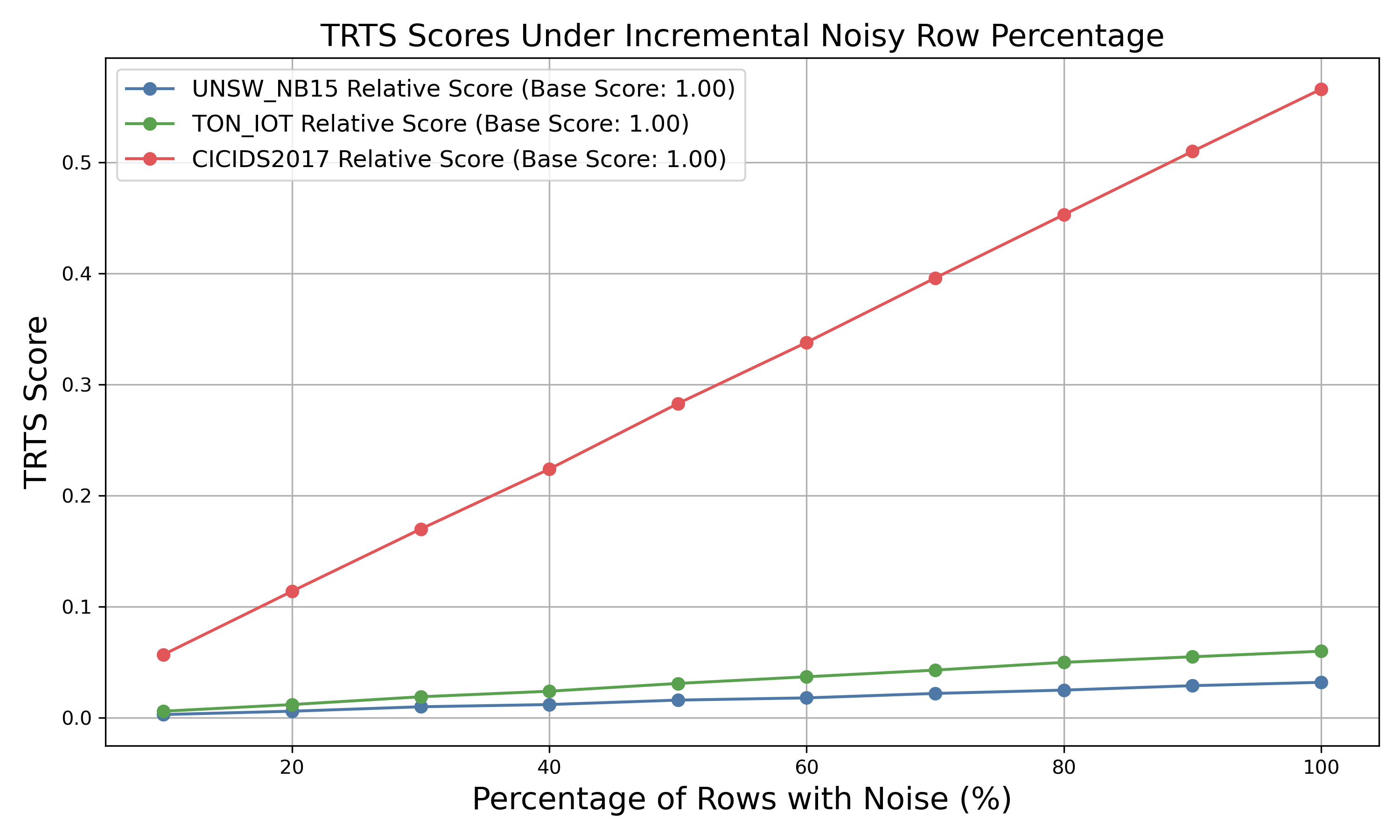}
    \end{subfigure}

    \begin{subfigure}{0.48\columnwidth}
        \includegraphics[width=\linewidth]{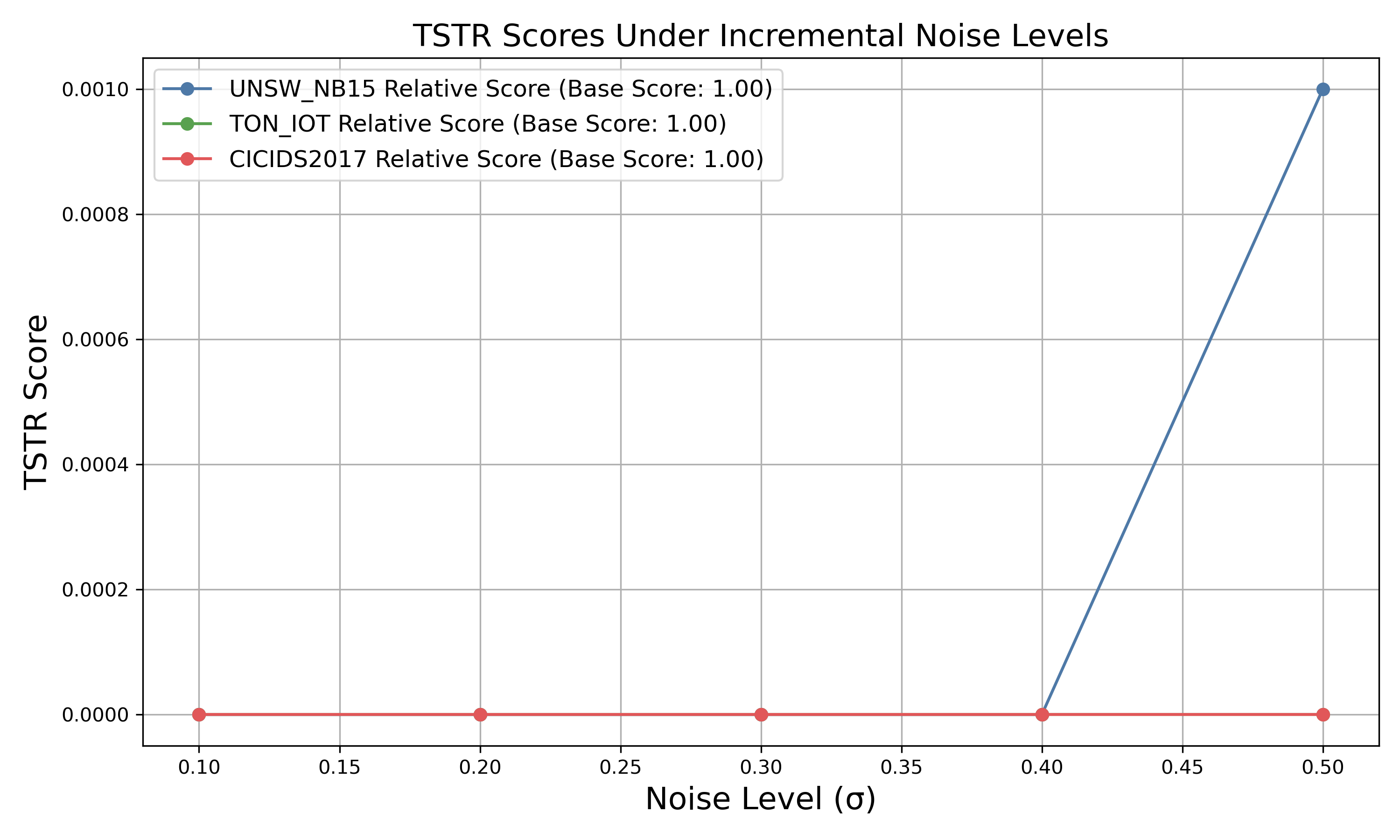}
    \end{subfigure}
    \begin{subfigure}{0.48\columnwidth}
        \includegraphics[width=\linewidth]{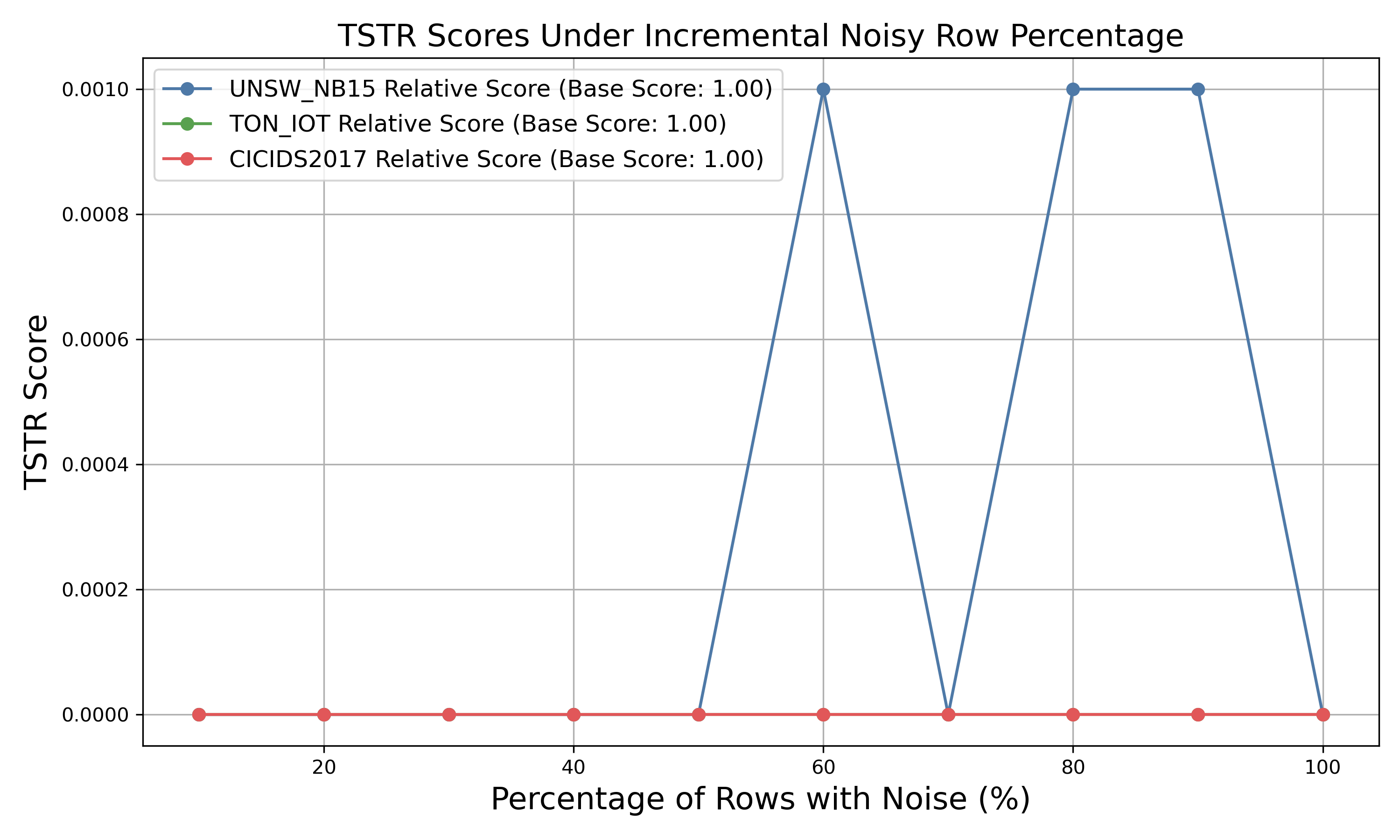}
    \end{subfigure}

    \caption{Results of the quality decrease experiment, illustrating the sensitivity of various evaluation metrics to noise in synthetic data. SDV Fidelity remains largely unaffected, indicating insensitivity to noise. FAED exhibits strong sensitivity, with scores deteriorating significantly as noise levels increase, effectively capturing structural disruptions. FPCAD shows a more gradual decline, but its behavior varies across datasets. RFIS scores increase with noise, reflecting reduced classification confidence and diversity. TRTS detects noise at higher levels but remains less effective than RFIS, while TSTR fails to detect noise altogether. These results highlight the varying effectiveness of metrics in assessing generative model robustness against noise-induced degradation.}
    \label{fig:res_quality_decrease}
\end{figure}

Figure \ref{fig:res_quality_decrease} summarize the results of the quality decrease experiments.
The findings highlight varying levels of sensitivity among different metrics in detecting noise in synthetic data. \\
\textbf{SDV Fidelity} remains largely unchanged across noise levels and datasets, indicating insensitivity to noise. For instance, in the UNSW-NB15 dataset, the relative score varies only slightly from 0.050 ($\alpha$ = 0.1) to 0.059 ($\alpha$ = 0.5), a trend consistent across all datasets. This suggests that SDV Fidelity fails to capture the impact of noise on data quality.\\
\textbf{FAED}, however, exhibits strong sensitivity to noise, with scores degrading significantly as noise levels increase. In the UNSW-NB15 dataset, the relative score plummets from -5.967 ($\alpha$ = 0.1) to -184.376 ($\alpha$ = 0.5), reflecting the metric’s ability to detect structural disruptions in the data. Similarly, FAED scores decline sharply across increasing noisy row percentages, confirming its robustness in capturing noise-induced distortions.\\
\textbf{FPCAD} shows a more gradual decline compared to FAED. In the UNSW-NB15 dataset, scores decrease from 0 ($\alpha$ = 0.1) to -0.022 ($\alpha$ = 0.5), with a similar trend across noisy row percentages. Interestingly, CICIDS-2017 exhibits an unexpected increase in FPCAD scores with noise, suggesting dataset-dependent limitations in its applicability.\\
\textbf{RFIS} scores increase as noise levels rise, indicating declining data quality. In the UNSW-NB15 dataset, the relative score rises from 0.565 ($\alpha$ = 0.1) to 0.660 ($\alpha$ = 0.5). This trend, consistent across datasets, suggests that RFIS effectively captures noise-induced degradation by reflecting reduced classification confidence and diversity.\\
\textbf{TRTS} detects noise at higher levels, with scores increasing modestly. In UNSW-NB15, the score rises from 0.000 ($\alpha$ = 0.1) to 0.033 ($\alpha$ = 0.5), while CICIDS-2017 shows a more pronounced increase from 0.135 to 0.566. However, TRTS remains less effective than RFIS, which captures both data quality and diversity more reliably.\\
\textbf{TSTR} fails to detect noise, with accuracy scores remaining at 1 and relative scores at 0 across all conditions. This suggests that noise either has minimal impact on model learning or that TSTR is inherently insensitive to data corruption.\\
In summary, the experiment highlights key differences in metric sensitivity. While SDV Fidelity remains largely unaffected by noise, FAED proves highly effective in detecting structural disruptions. RFIS also performs well, capturing noise-related quality degradation. However, FPCAD yields inconsistent results, particularly in dataset-specific cases. TRTS exhibit moderate noise sensitivity, while TSTR fails to detect noise altogether. These findings underscore the need for robust evaluation metrics tailored for generative models in tabular data.

\subsection{Experiment2: Mode Drop}





\begin{table*}[h!]
    \centering
    \captionsetup{width=\linewidth}
    \caption{Results of the mode drop experiment, summarizing the sensitivity of various metrics to mode drop across three datasets. SDV Fidelity shows limited sensitivity, detecting single mode drops in some cases but failing to capture significant degradation, especially in CICIDS-2017. FAED consistently detects mode drop, with significant score reductions for both single and successive mode drops, proving its robustness. FPCAD detects single mode drop but is inconsistent in capturing successive drops, particularly in CICIDS-2017. RFIS exhibits mixed results, detecting single mode drops but failing in extreme cases. TRTS is ineffective, with scores remaining unchanged across all datasets. TSTR shows moderate sensitivity, particularly in UNSW NB15 and CICIDS-2017. 
    The arrows ($\uparrow$ and $\downarrow$) indicate in which direction the relative score gets worse.}
    \label{tab:res_mode_drop}
    \makebox[\textwidth][c]{       
    \resizebox{\textwidth}{!}{
    \begin{tabular}{llrrr|rrrrr|rrrrr}
    \toprule
    \textbf{Metric} & \textbf{Dataset} & \textbf{Base Score} & \multicolumn{12}{c}{Relative Score} \\
    \cmidrule(lr){4-15}
    &       & &    \multicolumn{2}{c}{w/o 0/1 (Single)} & 
                \multicolumn{5}{c}{w/o Top N Combinations (Successive)} & \multicolumn{5}{c}{with Bottom \% Combinations (Extreme)} \\
    \cmidrule(lr){4-15}
     & & & 0 & 1 & 1 & 2 & 3 & 4 & 5 & 10 & 20 & 30 & 40 & 50 \\
    \midrule
    \multirow{3}{*}{SDV Fidelity $\uparrow$} & UNSW-NB15 & 0.983 & 0.028 & 0.052 & 0.026 & 0.014 & 0.010 & 0.015 & 0.012 & 0.101 & 0.107 & 0.099 & 0.101 & 0.095 \\
                                  & TON-IOT & 0.999 & 0.004 & 0.012 & 0.002 & 0.002 & 0.002 & 0.002 & 0.002 & 0.013 & 0.013 & 0.014 & 0.014 & 0.015 \\
                                  & CICIDS-2017 & 0.999 & 0.003 & 0.003 & 0.002 & 0.002 & 0.003 & 0.003 & 0.003 & 0.006 & 0.007 & 0.010 & 0.009 & 0.009 \\
    \midrule
    \multirow{3}{*}{FAED $\downarrow$} & UNSW-NB15 & 2.904 & -8.289 & -21.241 & -7.202 & -8.798 & -9.817 & -11.264 & -11.817 & -243.038 & -111.273 & -98.433 & -70.921 & -60.503 \\
                          & TON-IOT & 5.276 & -1.381 & -31.670 & -0.968 & -1.660 & -2.246 & -3.116 & -3.645 & -441.132 & -367.759 & -281.193 & -212.175 & -103.346 \\
                          & CICIDS-2017 & 1.349 & -39.966 & -17.418 & -3.222 & -17.076 & -19.750 & -20.683 & -22.894 & -54.742 & -66.356 & -44.836 & -42.613 & -38.046 \\
    \midrule
    \multirow{3}{*}{FPCAD $\downarrow$} & UNSW-NB15 & 0.438 & -1.032 & -5.962 & -4.907 & -3.364 & -3.059 & -3.811 & -3.200 & -27.025 & -11.318 & -9.780 & -8.012 & -5.741 \\
                           & TON-IOT & 0.188 & -1.375 & -9.514 & -0.107 & -0.209 & -0.273 & -0.405 & -0.494 & -128.737 & -107.203 & -80.955 & -70.437 & -32.095 \\
                           & CICIDS-2017 & 0.964 & -7.260 & -0.355 & 0.474 & 0.363 & -0.617 & -0.872 & -0.848 & -1.940 & -1.874 & -1.419 & -1.071 & -0.918 \\
    \midrule
    \multirow{3}{*}{RFIS $\uparrow$} & UNSW-NB15 & 1.882 & 0.870 & 0.855 & 0.023 & 0.026 & 0.055 & 0.101 & 0.061 & 0.683 & 0.785 & 0.814 & 0.830 & 0.831 \\
                          & TON-IOT & 1.728 & 0.727 & 0.724 & -0.023 & -0.046 & 0.008 & -0.010 & -0.027 & -0.198 & -0.189 & -0.176 & -0.087 & -0.008 \\
                          & CICIDS-2017 & 1.980 & 0.979 & 0.979 & 0.046 & 0.965 & 0.961 & 0.959 & 0.957 & 0.978 & 0.978 & 0.978 & 0.978 & 0.978 \\
    \midrule
    \multirow{3}{*}{TRTS $\uparrow$} & UNSW-NB15 & 1.000 & 0.000 & 0.000 & 0.000 & 0.000 & 0.000 & 0.000 & 0.000 & 0.000 & 0.000 & 0.000 & 0.000 & 0.000 \\
                           & TON-IOT & 1.000 & 0.000 & 0.000 & 0.000 & 0.000 & 0.000 & 0.000 & 0.000 & 0.000 & 0.000 & 0.000 & 0.000 & 0.000 \\
                           & CICIDS-2017 & 1.000 & 0.000 & 0.000 & 0.000 & 0.000 & 0.000 & 0.000 & 0.000 & 0.000 & 0.000 & 0.000 & 0.000 & 0.000 \\
    \midrule
    \multirow{3}{*}{TSTR $\uparrow$} & 
                            UNSW-NB15 & 1.000 & 0.319 & 0.681 & 0.000 & 0.000 & 0.000 & 0.000 & 0.000 & 0.094 & 0.097 & 0.078 & 0.088 & 0.078 \\
                            & TON-IOT & 1.000 & 0.237 & 0.763 & 0.000 & 0.000 & 0.000 & 0.000 & 0.000 & 0.002 & 0.001 & 0.001 & 0.000 & 0.000 \\
                            & CICIDS-2017 & 1.000 & 0.433 & 0.567 & 0.308 & 0.310 & 0.317 & 0.315 & 0.310 & 0.567 & 0.567 & 0.567 & 0.567 & 0.567 \\
    \bottomrule
    \end{tabular}}
    }
\end{table*}

This section analyzes how different metrics detect mode drop across the three datasets, with results summarized in Table \ref{tab:res_mode_drop}.\\ 
\textbf{SDV} Fidelity exhibits limited sensitivity to mode drop. While it detects single mode drops in UNSW-NB15, its impact on TON-IOT and CICIDS-2017 is negligible. Even under extreme mode drop conditions where 90\% of the most common values are removed—the CICIDS-2017 dataset experiences only a 0.6\% change, highlighting the metric`s inability to capture data degradation effectively.\\
\textbf{FAED}, in contrast, consistently detects mode drop. For single mode drop, removing labels 0 and 1 in UNSW-NB15 leads to relative score reductions of -9.223 and -20.067, respectively, with similar trends across datasets. Successive mode drops cause linear declines in FAED scores, indicating its robustness in capturing gradual data degradation. Extreme mode drop results further validate this, as scores progressively approach zero as lower frequency values are retained. \\
\textbf{FPCAD} effectively detects single mode drop, with scores dropping by -1.032 (label 0) and -7.515 (label 1) in UNSW-NB15. However, it fails to consistently capture successive mode drops, particularly in CICIDS-2017, where scores paradoxically increase. This inconsistency underscores its limitations in detecting structured data loss. Despite this, FPCAD remains effective under extreme mode drop conditions, though dataset diversity challenges its PCA-based approach. \\
\textbf{RFIS} provides mixed results. While it detects single mode drop (e.g., scores increasing to 0.870 and 0.855 for dropped labels in UNSW-NB15), successive mode drop results fluctuate, and TON-IOT even shows improvement. This highlights a limitation in Random Forest`s ability to accurately capture mode drop. Similarly, extreme mode drop results are inconsistent, with RFIS failing in TON-IOT while other datasets show an unexpected score increase, contradicting expectations. \\
\textbf{TRTS} is ineffective in detecting mode drop, maintaining an accuracy score of 1 and a relative score of 0 across all datasets. This suggests its reliance on accuracy limits its ability to recognize missing modes. \\
\textbf{TSTR}, however, shows better sensitivity. In UNSW-NB15, removing samples of labels 0 and 1 leads to relative scores of 0.319 and 0.681, respectively. However, for successive mode drops, scores remain stable except in CICIDS-2017. Under extreme mode drop conditions, the first two datasets show minor declines, while CICIDS-2017 experiences a significant drop, suggesting that the Random Forest classifier struggles when encountering missing modes not seen during training. \\
In summary, the results demonstrate that FAED is the most effective metric for detecting mode drop, capturing both gradual and extreme degradation. FPCAD and RFIS show partial effectiveness but lack consistency, particularly across datasets. SDV Fidelity and TRTS fail to provide meaningful insights, while TSTR and Synthesis Scores exhibit moderate detection capabilities. These findings highlight the importance of selecting robust evaluation metrics for generative models handling mode drop scenarios.

\subsection{Mode Collapse}
\begin{table}[h!]
    \centering
    \captionsetup{width=\linewidth}
    \caption{Results of the mode collapse experiment, highlighting the sensitivity of various metrics to mode collapse across different datasets. SDV Fidelity detects mode collapse in UNSW NB15 but shows limited effectiveness in more complex datasets like TON IOT and CICIDS-2017. FAED is highly sensitive, exhibiting significant score drops across all datasets, with particularly large declines in CICIDS-2017. FPCAD detects mode collapse but with lower sensitivity than FAED, showing consistent but less dramatic drops. RFIS displays significant score changes across datasets, unaffected by label splits. TRTS fails to detect mode collapse, except in the no-split category in CICIDS-2017. TSTR performs better with label splits, effectively detecting mode collapse. 
    The arrows ($\uparrow$ and $\downarrow$) indicate in which direction the relative score gets worse.}
    \label{tab:res_mode_collapse}
    \makebox[0.5 \textwidth][l]{       
    \resizebox{0.48\textwidth}{!}{
    \begin{tabular}{llrrrr}
    \toprule
    \textbf{Metric} & \textbf{Dataset} & \textbf{Base Score} & \multicolumn{3}{c}{Relative Score} \\
    \cmidrule(lr){4-6}
     & & & w/o 0 & w/o 1 & No Split \\
    \midrule
    \multirow{3}{*}{SDV Fidelity $\uparrow$} & UNSW-NB15 & 0.983 & 0.260 & 0.260 & 0.257 \\
                                  & TON-IOT & 0.999 & 0.022 & 0.022 & 0.022 \\
                                  & CICIDS-2017 & 0.999 & 0.019 & 0.019 & 0.019 \\
    \midrule
    \multirow{3}{*}{FAED $\downarrow$} & UNSW-NB15 & 2.904 & -50.175 & -50.223 & -43.060 \\
                          & TON-IOT & 5.276 & -10.477 & -14.333 & -9.260 \\
                          & CICIDS-2017 & 1.349 & -61.563 & -71.301 & -52.451 \\
    \midrule
    \multirow{3}{*}{FPCAD $\downarrow$} & UNSW-NB15 & 0.438 & -15.684 & -20.766 & -13.707 \\
                           & TON-IOT & 0.188 & -5.725 & -7.289 & -3.305 \\
                           & CICIDS-2017 & 0.964 & -19.804 & -19.815 & -19.763 \\
    \midrule
    \multirow{3}{*}{RFIS $\uparrow$} & UNSW-NB15 & 1.882 & 0.882 & 0.882 & 0.882 \\
                          & TON-IOT & 1.728 & 0.728 & 0.728 & 0.728 \\
                          & CICIDS-2017 & 1.980 & 0.980 & 0.980 & 0.980 \\
    \midrule
    \multirow{3}{*}{TRTS $\uparrow$} & UNSW-NB15 & 1.000 & 0.000 & 0.000 & 0.082 \\
                           & TON-IOT & 1.000 & 0.000 & 0.000 & 0.000 \\
                           & CICIDS-2017 & 1.000 & 0.000 & 0.000 & 0.499 \\
    \midrule
    \multirow{3}{*}{TSTR $\uparrow$} & UNSW-NB15 & 1.000 & 0.319 & 0.681 & 0.085 \\
                          & TON-IOT & 1.000 & 0.237 & 0.763 & 0.000 \\
                          & CICIDS-2017 & 1.000 & 0.433 & 0.567 & 0.000 \\
    \bottomrule
    \end{tabular}}}
\end{table}
The final robustness experiment examines mode collapse, with results summarized in Table \ref{tab:res_mode_collapse}. These results highlight how each metric detects mode collapse across different datasets.\\
\textbf{SDV Fidelity} detects mode collapse in UNSW-NB15 with minimal variation between collapse types. However, its significance diminishes in the more complex TON-IOT and CICIDS-2017 datasets, limiting its effectiveness.\\
\textbf{FAED} is highly sensitive to mode collapse, showing substantial drops across all datasets. For CICIDS-2017, the score declines by -61.563 and -71.301 for mode collapse in the ``w/o 0" and ``w/o 1" categories, respectively. The metric performs slightly better when mode collapse is applied without a label split, suggesting that splitting modes by labels introduces additional deviations from true values.\\
\textbf{FPCAD} also detects mode collapse but with lower sensitivity than FAED. In UNSW-NB15, the scores drop by -15.684, -20.766, and -13.707 across the different collapse types, maintaining consistency across datasets.\\
\textbf{RFIS} shows significant score changes due to mode collapse, with results remaining unaffected by label splits. This trend holds across all datasets, reinforcing its robustness in detecting collapse.\\
\textbf{TRTS} generally fails to detect mode collapse, except for the no-split category in CICIDS-2017, where accuracy drops by nearly half, yielding a relative score of 0.499.\\
\textbf{TSTR}, in contrast, performs better with label splits, detecting mode collapse more effectively than in the no-split condition, where accuracy remains relatively unchanged.\\
In summary, among the evaluated metrics, FAED is the most reliable in detecting mode collapse, followed by FPCAD and RFIS. SDV Fidelity struggles with complex datasets, while TRTS show limited reliability. TSTR performs best when labels are split, and Synthesis Scores confirm the loss of diversity due to collapse. These findings emphasize the need for robust metrics in detecting structural distortions in generative models.

\section{Conclusion and Future Work}
In this study, we tackled the challenge of evaluating generative models for tabular data by proposing three novel evaluation metrics: FAED, FPCAD, and RFIS. Our extensive empirical analysis, conducted on benchmark network intrusion detection datasets, revealed significant shortcomings in existing evaluation metrics such as Fidelity, Utility, TSTR, and TRTS, which fail to capture key generative modeling challenges, including quality degradation, mode drop, and mode collapse. Among the proposed novel metrics, FAED emerged as the most robust, consistently detecting structural distortions in generated data across various degradation scenarios. Its ability to identify noise-induced quality deterioration, mode drop, and mode collapse highlights its effectiveness as a general-purpose evaluation metric for tabular generative models. FPCAD demonstrated potential in certain cases but exhibited inconsistencies across datasets, suggesting that further refinement is needed to improve its stability and applicability. RFIS effectively captured data quality degradation and mode collapse but showed variability in detecting mode drop, emphasizing the need for dataset-specific considerations when deploying it as an evaluation metric. These findings underscore the necessity for more comprehensive and reliable evaluation frameworks tailored for generative models in tabular data settings. While traditional evaluation methods often overlook structural distortions in synthetic data, our proposed metrics offer a more nuanced and rigorous assessment of model performance.

Moving forward, further research is needed to refine FPCAD and enhance its robustness across diverse datasets. Additionally, the applicability of FAED and RFIS should be explored beyond network intrusion detection datasets, extending their use to broader domains such as financial fraud detection, healthcare analytics, and synthetic data generation for privacy-preserving applications. Another promising direction involves integrating these metrics into automated model selection pipelines, enabling more efficient evaluation and comparison of generative models without extensive manual intervention. Overall, our work contributes to the advancement of generative model evaluation for tabular data, establishing a structured methodology to quantify model quality and reliability. By addressing the limitations of existing approaches, we provide a foundation for more rigorous and systematic assessment practices, ultimately supporting the development of more robust and generalizable generative models for tabular data applications.

\bibliographystyle{IEEEtran}
\bibliography{reference}

\end{document}